\pdfoutput=1

\documentclass[11pt]{article}

\usepackage{hyperref}

\usepackage{ACL}

\usepackage{times}
\usepackage{latexsym}

\usepackage[T1]{fontenc}

\usepackage[utf8]{inputenc}

\usepackage{microtype}

\usepackage{inconsolata}
\usepackage{enumitem}
\usepackage{adjustbox}
\usepackage{graphicx}

\usepackage{pgfplots}

\usepackage{algorithm} 
\usepackage{algorithmic}
\usepackage{booktabs}
\usepackage{amsmath}
\usepackage{pifont}
\usepackage{amssymb} 

\usepackage{colortbl} 
\usepackage{xcolor} 
\usepackage{enumitem}
\setlist[itemize]{itemsep=-4pt, topsep=2pt}  

\usepackage{titlesec}
\usepackage[absolute,overlay]{textpos}

\pgfplotsset{compat=1.18}

\begin{document}

\begin{textblock*}{5cm}(2cm, 2.2cm)
    \includegraphics[width=1.2cm]{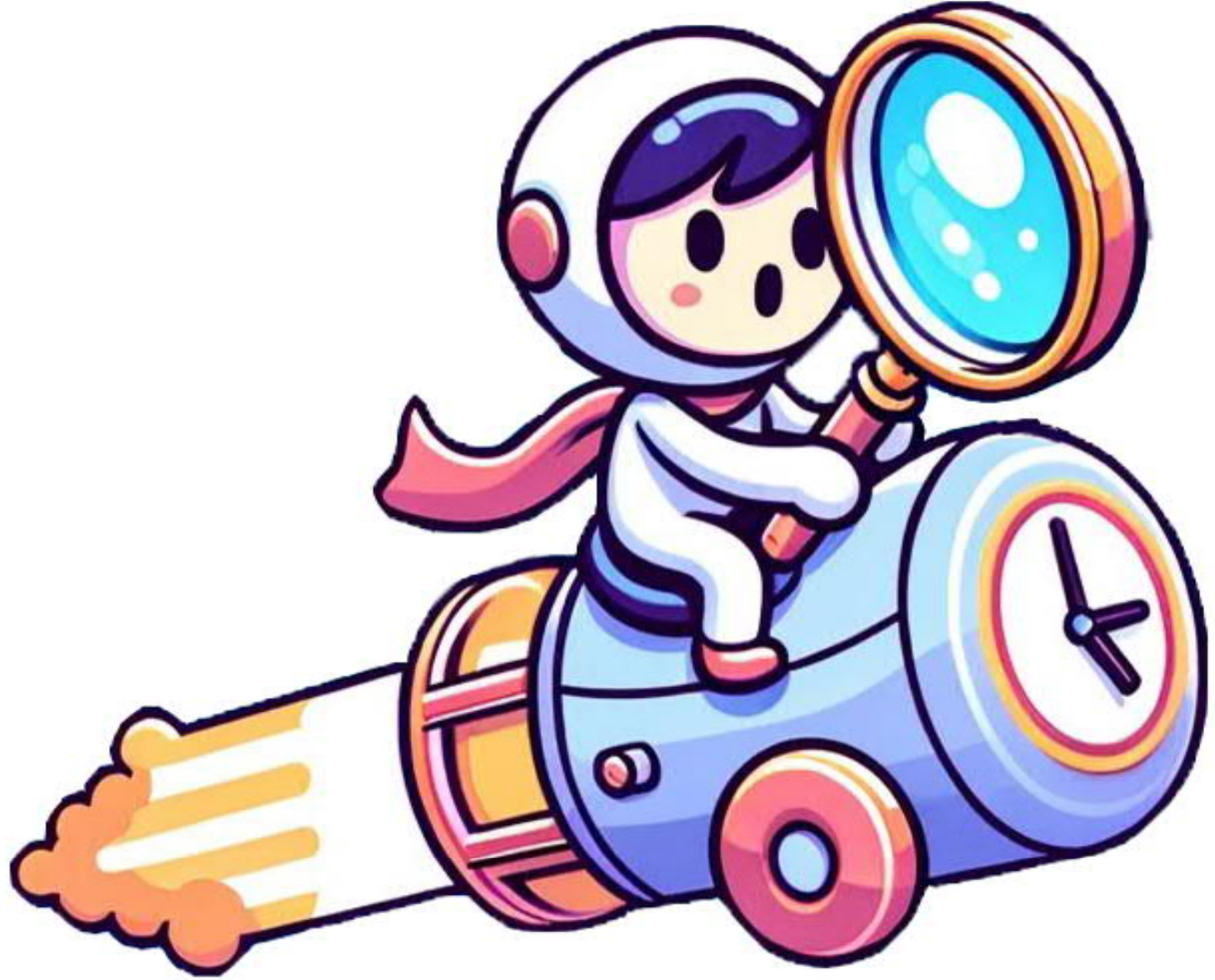}
\end{textblock*}

\title{\hspace{0.5cm}ONSEP: A Novel Online Neural-Symbolic Framework for Event Prediction Based on Large Language Model}

\author{
  Xuanqing Yu$^{1,2,3}$, Wangtao Sun$^{1,2,3}$,  Jingwei Li$^{1,2}$, Kang Liu$^{1,2}$,  Chengbao Liu$^{1,2 \dag}$, Jie Tan$^{1,2}$\\
  \textit{$^{1}$Institute of Automation, Chinese Academy of Sciences, Beijing, China} \\
  \textit{$^{2}$School of Artificial Intelligence, University of Chinese Academy of Sciences, Beijing, China} \\
  \textit{$^{3}$AI Lab, AIGility Cloud Innovation, Beijing, China} \\
  \texttt{\{yuxuanqing2021,sunwangtao2021,lijingwei2019,liuchengbao2016,jie.tan\}@ia.ac.cn} \\
  \texttt{\{kliu\}@nlpr.ia.ac.cn}
}
 



\setlength{\intextsep}{10pt plus 2pt minus 2pt}
\setlength{\textfloatsep}{3pt plus 2pt minus 2pt}

\titlespacing*{\section}{0pt}{4pt}{2pt}
\titlespacing*{\subsection}{0pt}{4pt}{4pt}
\titlespacing*{\subsubsection}{0pt}{4pt}{4pt}
\titlespacing*{\paragraph}{0pt}{2pt}{5pt}

\maketitle

\renewcommand{\thefootnote}{}  
\footnotetext{$^{\dag}$Corresponding author.}
\footnotetext{$^{*}$The data and code of this paper can be found at \url{https://github.com/aqSeabiscuit/ONSEP}.}

\begin{abstract}

In the realm of event prediction, temporal knowledge graph forecasting (TKGF) stands as a pivotal technique. Previous approaches face the challenges of not utilizing experience during testing and relying on a single short-term history, which limits adaptation to evolving data. In this paper, we introduce the Online Neural-Symbolic Event Prediction (ONSEP) framework, which innovates by integrating dynamic causal rule mining (DCRM) and dual history augmented generation (DHAG). DCRM dynamically constructs causal rules from real-time data, allowing for swift adaptation to new causal relationships. In parallel, DHAG merges short-term and long-term historical contexts, leveraging a bi-branch approach to enrich event prediction. Our framework demonstrates notable performance enhancements across diverse datasets, with significant Hit@k (k=1,3,10) improvements, showcasing its ability to augment large language models (LLMs) for event prediction without necessitating extensive retraining. The ONSEP framework not only advances the field of TKGF but also underscores the potential of neural-symbolic approaches in adapting to dynamic data environments.



\end{abstract}

\section{Introduction}

Event prediction is a widely researched topic \cite{zhao2021event, benzin2023survey} since accurate prediction of future events allows one to minimize losses associated with certain future events. To model large amounts of real-world event data that represent complex interactions between entities over time, the temporal knowledge graph (TKG) has been introduced \cite{ding2023forecasttkgquestions, yuan2023back}. TKG is used to represent structural relationships among entities through timestamped quadruples \((s, r, o, t)\), where \(s\) and \(o\) are entities, \(r\) is a binary relation between them, and \(t\) specifies the time when the event \((s, r, o)\) occurs. For example, the quadruple \((Angela\ Merkel, visit, China, 2014/07/04)\) indicates that Angela Merkel visited China on July 4, 2014. In this task, temporal knowledge graph forecasting (TKGF) aims to predict future events of the graph by inferring missing entities in a quadruple for a future time. This involves generating predictions for either the object entity \((s, r, ?, t + k)\) or the subject entity \((?, r, o, t + k)\) by utilizing historical data from previous snapshots, where \(k\) represents the number of time steps or intervals into the future beyond the current time \(t\). Existing research studies have provided theoretical methodologies for time-sensitive applications such as recommendation systems \cite{wang2022modeling, zhao2022time}, financial analysis \cite{li2023findkg}, and social crisis early warning systems \cite{gastinger2023dynamic}.

\begin{figure}[t]
    \centering
    \hspace*{-1.1cm} 
    \begin{adjustbox}{max width=1.26\columnwidth}
        \includegraphics[width=1.26\columnwidth]{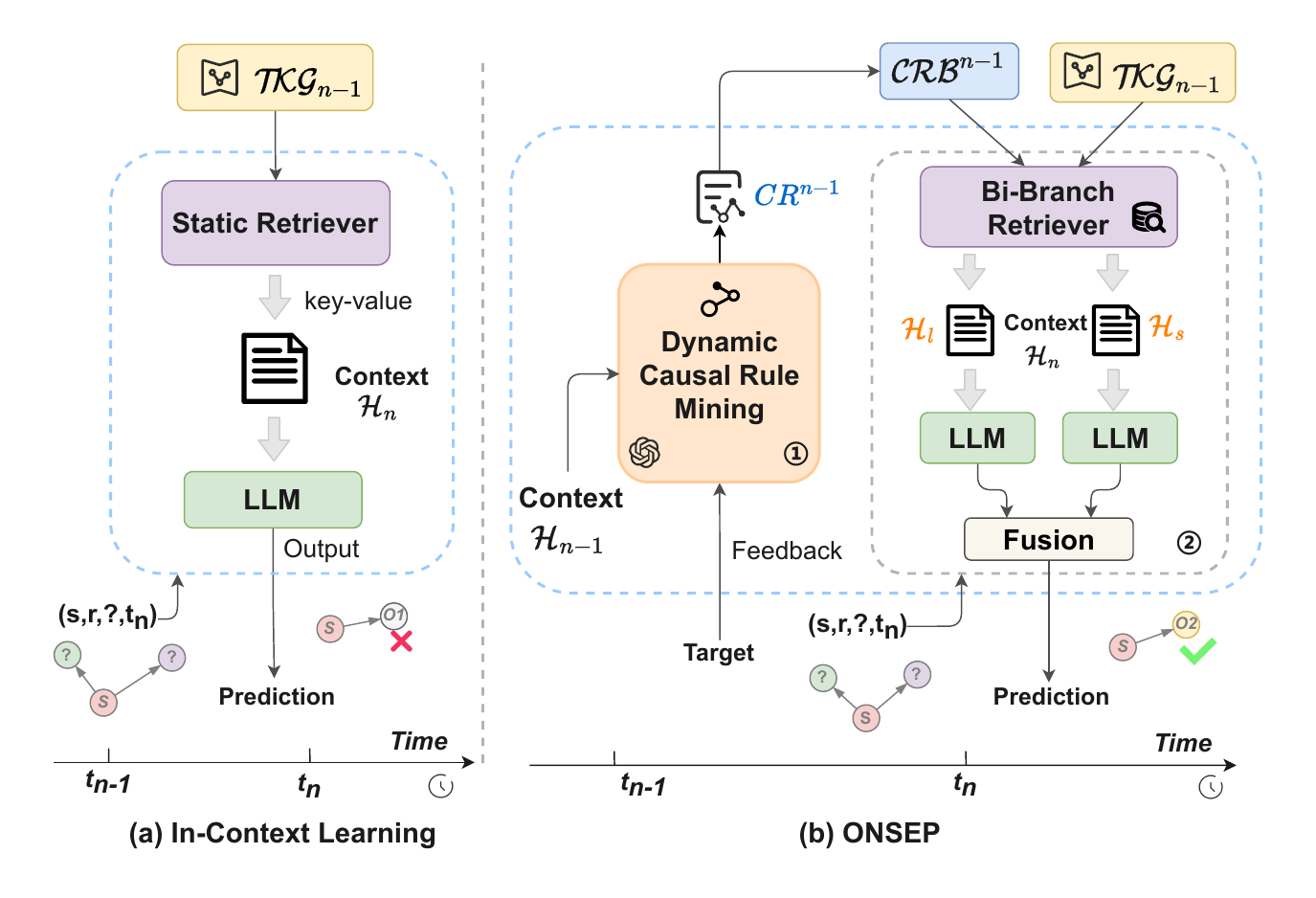}
    \end{adjustbox}
     \caption{Comparison of ONSEP and ICL Frameworks for Event Prediction with Schematic Overview of ONSEP's Core Components and Operational Processes.}
    \label{Figure1}
\end{figure}

Traditional approaches \cite{jin2019recurrent, zhu2021learning, li2021temporal, han2020xerte, sun2021timetraveler, li2022tirgn} involve converting event data into TKGs and combining graph neural networks (GNNs) and recurrent neural networks (RNNs) capture evolving entity relationships through embeddings. However, these methods must perform model training on specific datasets, which is resource-intensive. With the proven understanding and generative capabilities of large language models (LLMs), recent studies have explored methods on LLMs \cite{tao2023eveval, lee2023temporal, shi2023language}. The method proposed by \cite{lee2023temporal} offers a more adaptable method for TKGF with in-context learning (ICL) on LLMs, allowing LLMs to adapt to TKGF by using examples in the context, without fine-tuning.


Nevertheless, due to limitations in the length of history of LLM inputs, this approach may not fully capture long-term trends among events and cannot effectively leverage past insights, such as causal relationships between events. Imagine a TKG scenario that relates to everyday life. An example quadruple could be: `(\(Sarah\), \(Consult\), \(Dr.\ Smith\), \(2022/04/10\))', indicating that Sarah consulted Dr. Smith on April 10, 2022. In this case, the ICL method may use static key values like `\(Sarah\)' and  `\((Sarah, Consult)\)' to extract historical events. However, this method may not account for Sarah's tendency to begin consultations with phone calls months before meeting in person, This is evident in events such as `\(Discuss\ by\ telephone\)' being a preceding cause, for example, `(\(Sarah\), \(Discuss\ by\ telephone\), \(Dr.\ Smith, 2022/01/08\))'. This long-standing pattern of initiating consultations with a call is a crucial piece of historical context that short-term data analysis could miss. Without considering these longer-term causal interactions, the LLM-based method may not make accurate future predictions. Besides, in datasets like ICEWS \cite{Boschee2015ICEWSCoded}, the distribution of data containing entities and relations is dynamically changing. The emergence of new relations during test time poses challenges for traditional ICL methods, which rely on fixed keyword-key value matching and cannot adapt over time. This highlights the need for methods that can dynamically capture and apply updates in real-time, enhancing adaptability in event prediction.

To overcome these limitations, this paper proposes the \textbf{O}nline \textbf{N}eural-\textbf{S}ymbolic \textbf{E}vent \textbf{P}rediction (ONSEP) framework. It enhances both accuracy and adaptability in event prediction by addressing inadequate long-term causal relationship capture and enabling real-time adaptability for self-improvement without fine-tuning. Figure \ref{Figure1} illustrates a comparison of the ONSEP and ICL approaches. By contrast, ONSEP mainly has two novel components: 1) \textbf{\textit{Dynamic causal rule mining (DCRM)}}: Utilizing LLMs' external knowledge, DCRM semantically detects cause-effect links and dynamically constructs causal rules. This enables ONSEP to quickly adapt to new data and causal relationships, facilitating real-time updates and leveraging past experiences without extensive retraining. 2) \textbf{\textit{Dual History Augmented Generation (DHAG)}}: DHAG employs the \textit{long short-term bi-branch retriever (LSBBR)} and a \textit{hybrid model inference (HMI)} strategy to merge short-term and long-term historical contexts. The latter benefits from causal rules derived during the DCRM phase, enabling a broader event to be captured within the limits of historical input length. Inspired by the multi-branch fusion inference technology described in \cite{shi2023replug}, the HMI strategy applies weighted fusion to balance the contributions from both dual historical contexts. These innovations address the limitations of context length constraints in LLMs and improve the retrieval of relevant historical events over extended periods.

ONSEP shows significant performance gains on various datasets using InternLM2-7B model, achieving Hit@1 improvements over ICL of 9.63\%, 9.35\%, and 16.28\% at history length 100, and 8.14\%, 8.64\%, and 15.25\% at 200, and achieved competitive performance of embedding-based models trained on specific datasets. To summarize, our main contributions include:

\begin{itemize}[leftmargin=*,itemsep=1pt,topsep=1pt,parsep=0pt]
    \item We introduce DCRM, an innovative real-time adaptive causal learning module for LLMs that automatically updates the rule base at the snapshot level during testing.
    \item We develop the DHAG module, which uses LSBBR and HMI strategies, allowing LLMs to effectively use historical data from different time scales for causal analysis.
    \item Our framework includes an adaptive RAG solution that improves historical event retrieval and achieves self-improvement.
    \item We demonstrate ONSEP's effectiveness across various models and datasets, showcasing its ability to enhance black-box LLM inference without the need for fine-tuning or manual annotations, thereby providing a robust and adaptable solution for diverse event prediction tasks.
\end{itemize}

\section{Preliminaries}

\subsection{Temporal Knowledge Graph Forecasting}

A temporal knowledge graph (TKG) is structured as a time-sequenced series of multi-relational directed graphs. The TKG up to time \( t \) is represented as \( \mathcal{TKG}_t = \{\mathcal{G}_1, \mathcal{G}_2, \ldots, \mathcal{G}_t\} \), where each \( \mathcal{G}_t = (\mathcal{V}, \mathcal{R}, \mathcal{E}_t) \) represents a snapshot of the graph at time \( t \). Here, \( \mathcal{V} \) denotes the set of entities, \( \mathcal{R} \) the set of relations, and \( \mathcal{E}_t \) comprises timestamped facts as quadruples \( (s, r, o, t) \), with \( s, o \in \mathcal{V} \) and \( r \in \mathcal{R} \). TKGF aims to predict future states of the graph by inferring missing entities in a quadruple for a future time. This involves generating predictions for either the object entity \( (s, r, ?, t + k) \) or the subject entity \( (?, r, o, t + k) \) using historical data from previous snapshots \( \mathcal{TKG}_{t} \).

\subsection{ICL for Temporal Knowledge Graph Forecasting}

ICL enables LLMs to adjust to new tasks through contextual examples, without the need for fine-tuning. Specifically, in TKGF, ICL harnesses the adaptability of LLMs for forecasting by leveraging historical data. For a query of future event \( q=(s_q,r_q,?,t_n)\), where \( s_q \) is an entity and \( r_q \) is a relation at timestamp \( t_n \), this method employs static keys, such as entity \( s_q \) or the pair \( (s_q, r_q) \), to retrieve the historical event chain \( H_n(q) \), a set of quadruples, from previous snapshots \( \mathcal{TKG}_{n-1}=G_{1:n-1} \). Then the method constructs a prompt \( \theta_q \) based on \( H_n(q) \). The prediction \( y_q \) is generated by leveraging the LLM's probability distribution \( y_q \sim P_{\text{LLM}}(y_q | \theta_q) \), employing ICL to generate forecasts without further training.

To effectively handle multi-word entity and relation names, a numeric mapping \( \mathcal{M}_{\text{ent}} \) and \( \mathcal{M}_{\text{rel}} \) assigns unique labels to entities and relations. For example, candidate entities like [South Africa, China, New England] are mapped to numerical values [0, 1, 2] to align the outputs from the LLM with these candidates. During in-context learning, LLMs perform a forward pass to produce logits \( s \) for next-token predictions. These logits are then transformed into a probability distribution \( \boldsymbol{D} \) using the softmax function, representing the likelihood of each candidate being the target entity or relation.

\renewcommand{\thefootnote}{\arabic{footnote}}

\subsection{Causal Rules in TKG}

Causal rules\footnote{In this work, we use "causal" to intuitively convey the role these rules play in identifying and retrieving events with potential causative relations during real-time analysis, differing from the strict definition in statistical causal inference.} in TKG capture cause-and-effect links between events, denoted as \(CR(r_e, r_c): (X, r_e, Y, T_2) \leftarrow (X, r_c, Y, T_1)\), where \(X\) and \(Y\) represent anonymized entities, and \(T_1\) and \(T_2\) are timestamps, with \(T_1 < T_2\) ensuring the correct temporal sequence from cause to effect.

Extending this, we define a causal rule base (\(\mathcal{CRB}\)) as a set of tuples, each comprising a causal rule (CR) and its confidence score (\(\text{conf}\)). 
The \(\mathcal{CRB}\) for effect \(r_e\) at timestamp \(t_n\) is denoted as: \(\mathcal{CRB}^n(r_e) = \{ ((X, r_e, Y, T_2) \leftarrow (X, r_{c_i}, Y, T_1), \text{conf}^n_i) \mid 1 \leq i \leq m \} \), where \(r_{c_i}\) indicates the cause relation, \(r_e\) denotes the resulting relation influenced by \(r_c\), and \(\text{conf}^n_i\) is the confidence score for the \(i\)-th causal rule at timestamp \(t_n\), a real number within \([0, 1]\). Here, \(m\) represents the total number of causal rules considered in the rule base.

\section{Method}

\begin{figure*}[t]
    \centering
    \begin{adjustbox}{max width=\textwidth} 
    \includegraphics[width=1.0\textwidth]{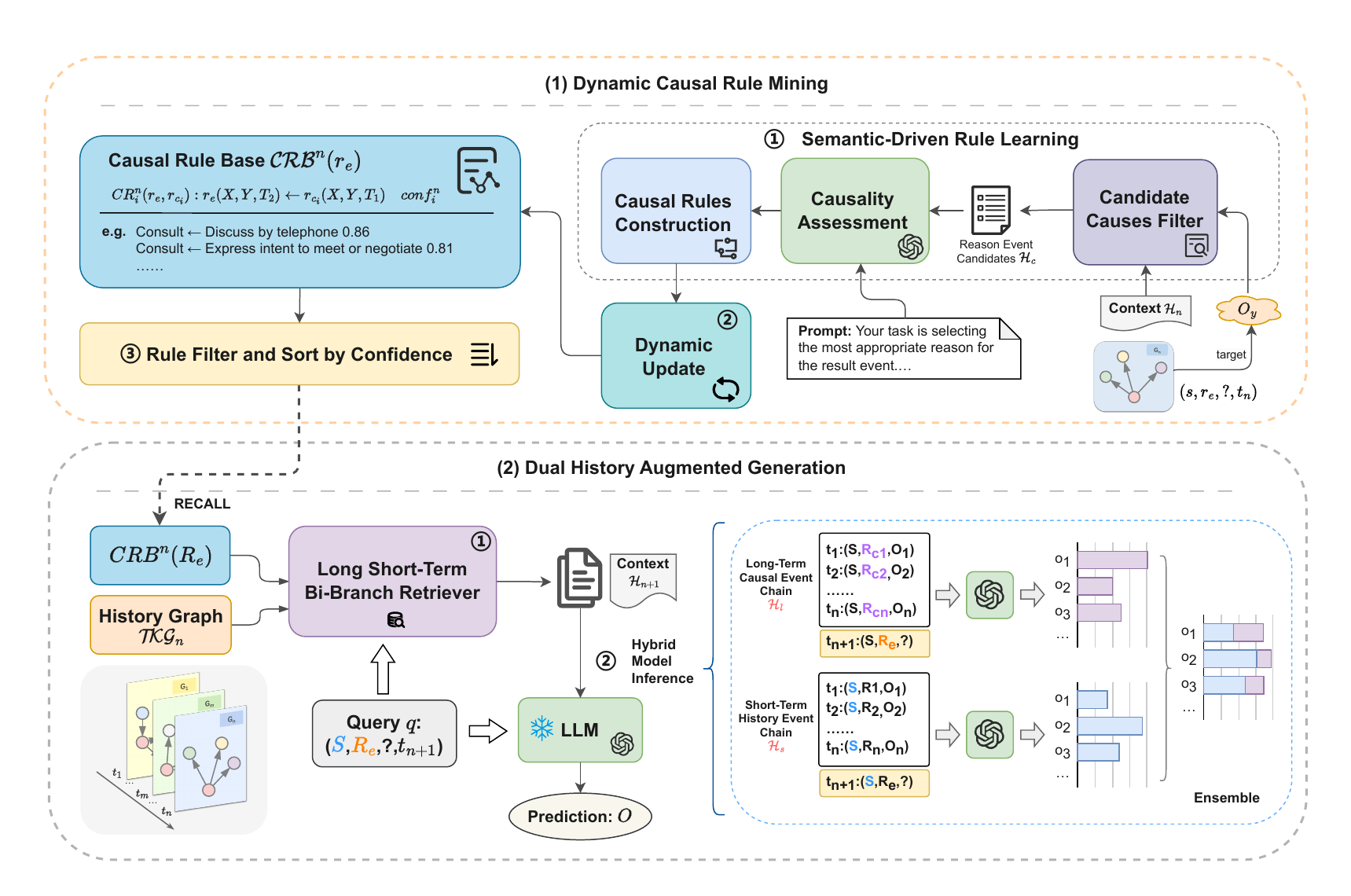}
    \end{adjustbox}
    \caption{Detailed structure of ONSEP framework with two phases: (1) Dynamic causal rule mining(§ \ref{DCRM}) and (2) dual history augmented generation(§ \ref{DHAG}). Specially, the DCRM phase employs a semantic-driven algorithm(§ \ref{csd-rl}) to identify causal rules and dynamically updates the rule base(§ \ref{DU}), incorporating a filtering and sorting mechanism(§ \ref{Filter&Sort}). (2) The DHAG phase utilizes a long short-term bi-branch retriever(§ \ref{LSTBBR}) alongside a hybrid model inference
    (§ \ref{HMI}) strategy to improve prediction accuracy.}
    \label{Figure2}
\end{figure*}

The ONSEP framework (Figure \ref{Figure2}) aims for event prediction via two key modules: (1) DCRM, which adaptively adjusts to changing data distributions during single-step prediction testing without requiring extensive training data, and (2) DHAG, which integrates patterns from short-term historical events with causality from long-term event developments. Detailed explanations are provided in the following sections.

\subsection{Dynamic Causal Rule Mining}
\label{DCRM}

The dynamic causal rule mining (DCRM) phase is crucial for identifying causal relationships within temporal knowledge graph. This phase utilizes a semantic-driven rule learning algorithm to discover causal rules. It is followed by a dynamic update module that updates the causal rule base (\(\mathcal{CRB}\)), which is closely followed by a mechanism for rule filtering and sorting by confidence.

\subsubsection{Semantic-Driven Rule Learning}
\label{csd-rl}

This module is designed for the reflective learning of causal rules, structured around three core steps: candidate causes filter, causality assessment, and causal rule construction. Initially, it retrieves historical context \(\mathcal{H}_n\) for all queries from \( \boldsymbol{G_n} \) and real-time feedback \(O_y\), which is the verified outcome for a query obtained when the system receives new data, to filter potential reason events. In the causality assessment phase, these candidates are then evaluated using a LLM to determine the most plausible cause event \(r_c\). Finally, the identified cause event, along with the query's effect event \(r_e\) forms the basis for constructing a causal rule \(CR(r_e, r_c)\). The algorithm of semantic-driven rule learning is outlined in Appendix \ref{Appendix:semantic_driven_rule_learning}. 

\paragraph{Candidate Event Filter}

In the candidate event filtering phase, we utilize historical context and real-time feedback related to the query to filter potential causal events. Given a query \( q = (S_q, R_e, ?, t_n) \) at time \( t_n \), we define the context \( \mathcal{H}_n(S_q) \) as a historical sequence of events associated with subject \( S_q \), expressed as \( \mathcal{H}_n = \{(S_q, r, o, t) \in \mathcal{G}_n\} \). We select candidates based on the criterion that the object \( o \) is the same as the target \( O_y \), thereby forming a set of candidate reason events \( \mathcal{H}_c \), formalized as:
\[
    \mathcal{H}_c(q) = \{(S, r_c, \hat{o}, \hat{t}) \in \mathcal{H}_n(S_q) \mid \hat{o} = O_y\}.
\]
Here, \( r_c \) represents a potential causal relation, \(\hat{o}\) is the filtered object that matches the target \( O_y \), and \(\hat{t}\) is the timestamp associated with the event. This methodology aims to pinpoint potential causative relations corresponding to \( r_e \) within the event sequence that links the query’s subject \( S \) to the target entity.

For each distinct causative event \( r_c \) within the candidates of reason events, we first extract the set of quadruples from \( \mathcal{H}_c \) that contain \( r_c \), which we denote as \( \mathcal{T}_{r_c} \): \[ \mathcal{T}_{r_c} = \{(s, r, o, t) \mid r = r_c, (s, r, o, t) \in \mathcal{H}_c(q)\}. \]
We then compute the support number \( supp(r_c) \) as the count of quadruples of this set: \( supp(r_c) = |\mathcal{T}_{r_c}|. \)
Subsequently, we calculate the coverage rate \( cove(r_c) \) for each event \( r_c \) using:
\begin{equation}
    cove(r_c) = \frac{supp(r_c)}{|\mathcal{H}_c(q)|}.
\end{equation}
This coverage rate aids in assessing the confidence level of causal rules and supports the review and selection of causal event candidates by the LLM.

\paragraph{Causality Assessment} 

\begin{table}[ht]
\centering
\resizebox{\columnwidth}{!}{%
    \begin{tabular}{p{1.0\columnwidth}}
    \toprule
    \textbf{LLM-based Causal Event Selection} \\
    \midrule
    Your task is selecting the most appropriate reason for the result event. The result event is \{ \textit{r\_e }\}. \\
    Below is a list of possible reasons: \\
    \{ \textit{candidate causal events: [label]. [candidate]} \} \\
    The most appropriate reason is: \\
    \bottomrule
    \end{tabular}
}
\caption{Causal Event Selection Prompt Template.}
\label{tab:cause_selection}
\end{table}

In this step, the causal evaluation utilizes the powerful semantic understanding capabilities of the LLM to assess the degree of causal association between a set of potentially causal events. This evaluation aims to select the causal rules that are most relevant to the current query and feedback goals.
A pivotal aspect of our methodology involves the formulation of a structured prompt \(\theta_1\) (Table \ref{tab:cause_selection}), designed to direct the LLM towards discerning the most pertinent causal link between a given result event (\(r_e\)) and potential causes within the candidate reason events set $\mathcal{H}_{c}(q)$. 
The selection mechanism utilizes LLM to obtain the logits \( \boldsymbol{L} \) for numerically mapped candidate reasons, as same as the way how to generate an output for each test query. These logits represent the preliminary evaluation of each candidate's likelihood to be the true causal event. To convert these logits into normalized probabilities, we apply the softmax function, yielding the probability \( p_{r_{c_i}} \) for each candidate reason \( r_{c_i} \), mapped to label $id_i$ using $\mathcal{M}_{rel}$, reprented as \( \boldsymbol{s} = LLM(\theta_1(r_e, \mathcal{H}_{c}(q)))\), where \(\boldsymbol{s}\) are the logits produced by LLM, The probability of each candidate reason \( r_{c_i} \) is then computed by:
\begin{equation}
    p(r_{c_i}) = \frac{e^{\boldsymbol{s}[id_i]}}{\sum\limits_{j \in \mathcal{M}_{\text{rel}}} e^{\boldsymbol{s}[id_j]}}.
\end{equation}

Based on these probabilities, the top-\(k\) candidates are selected. The confidence \( conf_{i}^{t} \) for each causal rule at timestamp \( t \) is determined by combining the probability \( p(r_{c_i}) \) with the coverage score \( cove(r_{c_i}) \), can obtained as:

\begin{equation}
    conf^{t}(r_{c_i}) = \alpha \cdot p(r_{c_i}) + (1 - \alpha) \cdot cove(r_{c_i})
\end{equation}

where \( \alpha \) is a tuning parameter that balances the contribution of the probability \( p(r_{c_i}) \) and the coverage score \( cove(r_{c_i}) \), integrating the causal assessment with historical occurrence insights.

\paragraph{Causal Rule Construction}

Following the causality assessment step, a set of selected causal events (\(r_c\)) and their corresponding confidence scores at the current time step (\(t_n\)) are obtained. These causal rules are formalized in a structured format, with each rule at timestamp $t$ represented as \(CR^{t}(r_e,r_c) = (X, r_e, Y, T_2) \leftarrow (X, r_c, Y, T_1) \), accompanied by a confidence value \(conf^{t}(CR(r_e,r_c))\), can be obtained by $conf^{t}(r_c).$

\subsubsection{Dynamic Update}
\label{DU}

The dynamic update module for the causal rule base adds new rules directly and updates existing ones by adjusting their confidence levels. It assigns a confidence score \(c_t\) to each causal rule \(CR\), linking a pair of relations, and this score is dynamically updated at each time step \(t\). This score merges the previous confidence \(c_{t-1}\) with the current confidence \(conf\) calculated at time \(t\), effectively updating the confidence for each pair of relations within the causal rule. 

It uses a smoothing factor (\(\theta\)) to stabilize confidence adjustments, and a growth factor (\(\beta\)) to incrementally evaluate the reliability of the circular verification rules, but the historical confidence cannot exceed 1. The update formula of the existing rules is:
\begin{equation}
    c_t = \theta \cdot f_g(c_{t-1}) + (1 - \theta) \cdot conf
\end{equation}
\begin{equation}
    f_g(c_{t-1}) = \min(c_{t-1} \cdot (1 + \beta), 1)
\end{equation}
where \(c_t\) is the updated confidence, \(conf\) the current evaluated confidence of \(CR\), and \(c_{t-1}\) the previous confidence, \(f_g\) represent the grow function with the growth factor \(\beta\).
 
This approach ensures continuous optimization of the rule base, allowing for adaptability in light of distribution changes over time.

\subsubsection{Rule Filter and Sort by Confidence}
\label{Filter&Sort}

In the final module of the DCRM, rules falling below a predefined confidence threshold, denoted as \(conf_{min}\), are filtered out to ensure the usability of the causal rule base. Subsequently, the remaining rules are sorted based on their confidence levels, allowing rules with higher confidence to take precedence in the subsequent reasoning phase.

\subsection{Dual History Augmented Generation}
\label{DHAG}

DHAG, an innovative RAG variant, introduces a long short-term bi-branch retriever (LSTBBR) coupled with a hybrid model inference (HMI) strategy. DHAG synergizes short-term historical event patterns with long-term causal trajectories, ensuring a comprehensive historical context. Unlike previous methods that model long and short-term histories through data associations, DHAG provides a more robust handling of low-frequency events, unaffected by irrelevant noise in long histories.

\subsubsection{Long Short-Term Bi-Branch Retriever}
\label{LSTBBR}

The long short-term bi-branch retriever (LSTBBR) within the DHAG framework incorporates a dual retrieval strategy to enhance the predictive model with a rich historical context, comprising both short-term and long-term historical events. For each prediction query \( q = (S_q, R_e, ?, t_{n+1}) \), LSTBBR extracts two distinct historical contexts: \( \mathcal{H}_s \) representing the short-term history event chain and \( \mathcal{H}_l \) representing the long-term causal event chain. This dual approach ensures a comprehensive understanding of the immediate events as well as the underlying long-term causal influences. The short-term history event chain (\( \mathcal{H}_s \)) focuses on capturing the most recent events that are temporally proximate to the prediction query, providing an immediate context that reflects the latest developments. To construct \( \mathcal{H}_s \), the system retrieves events from \( \mathcal{H}_n(S_q) \), sorting them by timestamp and truncating to include only the \( L \) most recent events. This truncated list of events forms \( \mathcal{H}_s \), which is directly associated with the query's subject \( S_q \), thereby providing a snapshot of the most immediate historical backdrop relevant to the query.

On the other hand, the long-term causal event chain (\( \mathcal{H}_l \)) is designed to uncover the broader causal dynamics that have shaped the subject \( S \) over a more extended period. This is achieved by initially retrieving cause rules from \( \mathcal{CRB}(r_e) \) using a RECALL mechanism, which employs a key-value approach where the effect relation \( r_e \) is used as the key to efficiently extract associated causal rules.

Subsequently, cause events are filtered to construct the long-term cause event chain \( \mathcal{H}_l \) from \( \mathcal{H}_n(S_q) \). The events in \( \mathcal{H}_l \) are determined by the criteria: \( (S_q, R_{c_i}, o_i, t_i) \) where \( CR(R_e, R_{c_i}) \in \mathcal{CRB}(R_e) \) and \( (S, R_{c_i}, o_i, t_i) \in \mathcal{H}_n(S_q) \), with \( t_i < t_{n+1} \). Similar to \( \mathcal{H}_s \), \( \mathcal{H}_l \) is truncated to include only the most recent \( L \) events since limited by the max length of model input.

\subsubsection{Hybrid Model Inference}
\label{HMI}

Before inference, a numerical label mapping technique preprocesses multi-word entities to prepare them for integration into the model, similar to the causality assessment in DCRM. The essence of the hybrid model inference (HMI) strategy involves merging the query \(q\) with the short-term history event chain \(H_s\) and the long-term causal event chain \(H_l\). This allows the LLM to produce distinct probabilities for each context, which are then combined using a weighted ensemble approach.

For the short-term context, the logits \(\boldsymbol{s}_1\) are obtained by \( \boldsymbol{s}_1 = p(y | q \oplus \mathcal{H}_s) \), where \( \oplus \) symbolizes concatenation. The logits \(\boldsymbol{s}_1\) are then normalized using the softmax function to produce a probability distribution \(\boldsymbol{D}_1 = \text{softmax}(\boldsymbol{s}_1)\). Similarly, for the long-term context, the logits \(\boldsymbol{s}_2\) are derived by \( \boldsymbol{s}_2 = p(y | q \oplus \mathcal{H}_l) \), and the corresponding probability distribution \(\boldsymbol{D}_2\) is obtained by applying the softmax function to \(\boldsymbol{s}_2\), resulting in \( \boldsymbol{D}_2 = \text{softmax}(\boldsymbol{s}_2) \). The integration strategy involves a weighted ensemble of $\boldsymbol{D}_1$ and $\boldsymbol{D}_2$, formulated as:
\begin{equation}
    \boldsymbol{D} = \boldsymbol{D}_1 \cdot (1 - \lambda) + \boldsymbol{D}_2 \cdot \lambda.
\end{equation}
Here, $\lambda$ is a tuning parameter that balances the contributions of short-term and long-term contexts. This unified distribution $\boldsymbol{D}$ ranks candidate entities by their relevance to $q$, leveraging the nuanced insights from both $\mathcal{H}_s$ and $\mathcal{H}_l$. By doing so, the HMI strategy enhances the LLM's accuracy in entity predictions, grounded in a comprehensive understanding of dual historical contexts.

\section{Experiments}
\subsection{Experimental Settings}
\paragraph{Datasets}
Our experimental evaluation is carried out on a subset of the integrated crisis early warning system (ICEWS) dataset, which includes versions such as ICEWS14 \cite{garcia2018learning}, ICEWS05-15 \cite{garcia2018learning}, and ICEWS18 \cite{jin2019recurrent}. These datasets are composed of timestamped records of political events, making them highly suitable for conducting temporal analysis. They are widely recognized as benchmark datasets on TKGF. Each event is represented as a tuple, such as (Barack Obama, visit, Malaysia, 2014/02/19), capturing diverse political activities across various time periods. 

\paragraph{Evaluation Metrics}

To evaluate our method's efficiency in ranking event candidates, we use link prediction metrics like Hit@k (where \(k=1, 3, 10\)). This measures the precision of our model in forecasting future events within the top \(k\) predictions. Higher Hit@k values indicate more accurate rankings, which, in product recommendations, translate to better purchase predictions and greater economic benefits.

\paragraph{Baselines}

We primarily compare ONSEP with the ICL method. Additionally, we have selected several traditional supervised models based on embedding methods for performance comparison, including RE-NET \cite{jin2019recurrent}, CyGNet \cite{zhu2021learning}, RE-GCN \cite{li2021temporal}, xERTE \cite{han2020xerte}, and TITer \cite{sun2021timetraveler} in TKG. We also compare our method with a variety of LLMs. 

Further information about the datasets, evaluation metrics, LLMs used and implementation specifics can be found in the Appendix \ref{Appendix:experimental_details}.

\begin{table*}[htbp]
\centering
\resizebox{\textwidth}{!}{%
\begin{tabular}{@{}ccccccccccc@{}}
\toprule
& & \multicolumn{3}{c}{ICEWS14} & \multicolumn{3}{c}{ICEWS05-15} & \multicolumn{3}{c}{ICEWS18} \\
\cmidrule(lr){3-5} \cmidrule(lr){6-8} \cmidrule(lr){9-11}
Model & Train & Hit@1 & Hit@3 & Hit@10 & Hit@1 & Hit@3 & Hit@10 & Hit@1 & Hit@3 & Hit@10 \\
\midrule
RE-NET & \ding{51}  & 0.301& 0.440& 0.582& 0.336& 0.488& 0.627& 0.197& 0.326& 0.485\\
CyGNet & \ding{51}  & 0.274& 0.426& 0.579& 0.294& 0.461& 0.616& 0.172& 0.310& \underline{0.469}\\
RE-GCN & \ding{51}  & 0.316& \textbf{0.472}& \textbf{0.617}& 0.373& \textbf{0.539}& \textbf{0.685}& \textbf{0.224}& \textbf{0.368}& \textbf{0.527}\\
xERTE & \ding{51}  & \underline{0.327}& 0.457& 0.573& \underline{0.378} & 0.523& 0.639& 0.210& \underline{0.335}& 0.465\\
TITer & \ding{51}  & \textbf{0.328}& \underline{0.465}& \underline{0.584}& \textbf{0.383}& \underline{0.528}& \underline{0.649}& \underline{0.221}& \underline{0.335}& 0.448\\
\midrule
\midrule
InternLM2-7B-ICL (L=100) & \ding{55} & 0.301 & 0.432 & 0.560 & 0.353& 0.507& 0.647& 0.172 & 0.289 & 0.434 \\
InternLM2-7B-ONSEP (L=100) & \ding{55} & \underline{0.330}& \underline{0.464}& \underline{0.570}& \underline{0.386}& \underline{0.546}& \underline{0.662}& \underline{0.200}& \underline{0.324}& 0.443\\
\rowcolor{gray!30}
\textit{$\Delta$ Improve} & & 9.63\%& 7.41\%& 1.79\%& 9.35\%& 7.69\%& 2.32\%& 16.28\%& 12.11\%& 2.07\%\\
\midrule
InternLM2-7B-ICL (L=200) & \ding{55} & 0.307 & 0.443 & 0.567 & 0.359& 0.520& 0.659& 0.177 & 0.300 & \underline{0.447} \\
InternLM2-7B-ONSEP (L=200) & \ding{55} & \textbf{0.332}& \textbf{0.465}& \textbf{0.577}& \textbf{0.390}& \textbf{0.551}& \textbf{0.668}& \textbf{0.204}& \textbf{0.333}& \textbf{0.453}\\
\rowcolor{gray!30}
\textit{$\Delta$ Improve} & & 8.14\%& 4.97\%& 1.76\%& 8.64\%& 5.96\%& 1.37\%& 15.25\%& 11.00\%& 1.34\%\\
\bottomrule
\end{tabular}
}
\caption{Performance comparison among LLM-based ICL and traditional embedding-based methods on three datasets with time-aware metrics (Hit@k). The highest performance is highlighted in \textbf{bold}. \textit{$\Delta$ Improve} represents the percentage improvement of ONSEP over ICL. The results of the embedding-based models are excerpted from \cite{li2022tirgn}.}

\label{tab:performance_comparison}
\end{table*}

\subsection{Performance Comparison}

To evaluate whether ONSEP surpasses the previous best event prediction method based on LLMs, ICL, we conduct experiments with historical inputs of 100 and 200, using InternLM2-7B for all methods. As Table \ref{tab:performance_comparison} shows, our method outperforms ICL across all three datasets. Specifically, with a history length of 100, ONSEP achieves Hit@1 improvements of 9.63\%, 9.35\%, and 16.28\%, and with 200, the gains are 8.14\%, 8.64\%, and 15.25\%. These results confirm ONSEP's capability to exceed the performance of the previous ICL method. These results underscore ONSEP's superior performance over ICL, particularly in Hit@1 accuracy, indicating a notable enhancement in precision.

While ONSEP trails behind some trained embedding models in Hit@10 due to the LLM's reliance on ranking candidate entities from the input history rather than all entities, it achieves top performance in Hit@1 for the ICEWS14 and ICEWS05-15 datasets. Additionally, it shows closely competitive results in other metrics. This highlights ONSEP's capability to significantly enhance accuracy. For a detailed exploration of ONSEP's operational dynamics and its real-world applicability, see the ICEWS14 case study in Appendix \ref{Appendix:case_study}.

\subsection{Inductive Setting and the Effectiveness of DCRM}
\begin{table}[h]
\centering
\resizebox{\columnwidth}{!}{%
\begin{tabular}{@{}l|>{\raggedright\arraybackslash}p{4cm}|ccc@{}}
\toprule
& Method (InternLM2-7B) & Hit@1 & Hit@3 & Hit@10 \\ \midrule
(i) & ICL & 0.156 & 0.265 & 0.382 \\ \midrule
(ii) & ONSEP & \textbf{0.186} & \textbf{0.305} &\textbf{ 0.424} \\ \midrule
(iii) & ONSEP - inductive  & \underline{0.184} & \underline{0.302} & \underline{0.422} \\ \midrule
(iv) & ONSEP w/o DCRM (ICL w/ DHAG) - inductive & 0.168 & 0.292 & 0.416 \\ \bottomrule
\end{tabular}
}
\caption{Analysis on DCRM module under inductive setting. Utilizing causal rules derived by ONSEP from the test graph \(G_{test_1}\) (ICEWS14, with a history input length of L=50) and applying them to a different test graph \(G_{test_2}\) (ICEWS18, with a history input length of L=30).}
\label{tab:Inductive Setting}
\end{table}

To assess the transferability of causal rules mined during test-time iterations, we conduct experiments where rules learned from ICEWS14 are applied to predictions on ICEWS18, with findings presented in Table \ref{tab:Inductive Setting}. There are significant temporal spans and differences in data distribution between ICEWS14 and ICEWS18. For example, ICEWS18 includes entities such as Donald Trump, who served as the President of the United States from 2017 to 2021, which are not present in ICEWS14.

With only a 20.3\% overlap (see Table \ref{table:overlap_statistics}) in entities between the two datasets, our inductive experimental setup demonstrates performance improvements. Comparing the ICL method without pre-loaded rules (i) to the approach using ICEWS14-derived rules (iv), we observe a significant performance boost in the latter, indicating that the DCRM-mined causal rules are generalizable across datasets with similar relational structures, thereby improving inference.

Further analysis on the DCRM module's impact shows that incorporating DCRM (iii) versus not incorporating it (iv) leads to enhanced performance across all metrics, including a notable 9.52\% increase in Hit@1. This underscores DCRM's effectiveness in improving the accuracy and recall of inference.

Interestingly, real-time rule mining with DCRM without pre-loaded rules (ii) is slightly higher than that of the inductive setting (iii) (i.e., with pre-loading the rules learned from \(G_{test_1}\)), suggesting that relying on potentially outdated rules may hinder adaptation to new data. Due to the smoothing setup in dynamic updates, the old rule set may introduce some interference, hindering rapid adaptation to the new test set. This emphasizes the need for dynamic rule updates to ensure model relevance and effectiveness across varying datasets with dynamically changing distributions.

\subsection{Effectiveness of DHAG}

To assess the DHAG module's impact, we explored how blending short-term and long-term history contexts affects performance on the ICEWS14 dataset, with similar findings observed on other datasets. We vary the Weighted Fusion Ratio from 0 to 1, as shown in Figure \ref{fig:ensemble_weight}, with details in Appendix \ref{Appendix:ensemble_weight}.

Our findings indicate a clear pattern: a \(\lambda\) value of 0, which effectively uses only short-term history chain akin to the baseline ICL method, leads to lower performance. Conversely, a \(\lambda\) of 1, relying solely on long-term reasoning, also underperforms compared to a balanced approach. The optimal performance at a fusion ratio of 0.1 indicates that the model primarily utilizes short-term dynamic history for reasoning, while also incorporating long-term causal knowledge acquired during test time to improve its effectiveness.

\begin{figure}[t]
    \centering
    \resizebox{\columnwidth}{!}{%
    \includegraphics[width=1.0\textwidth]{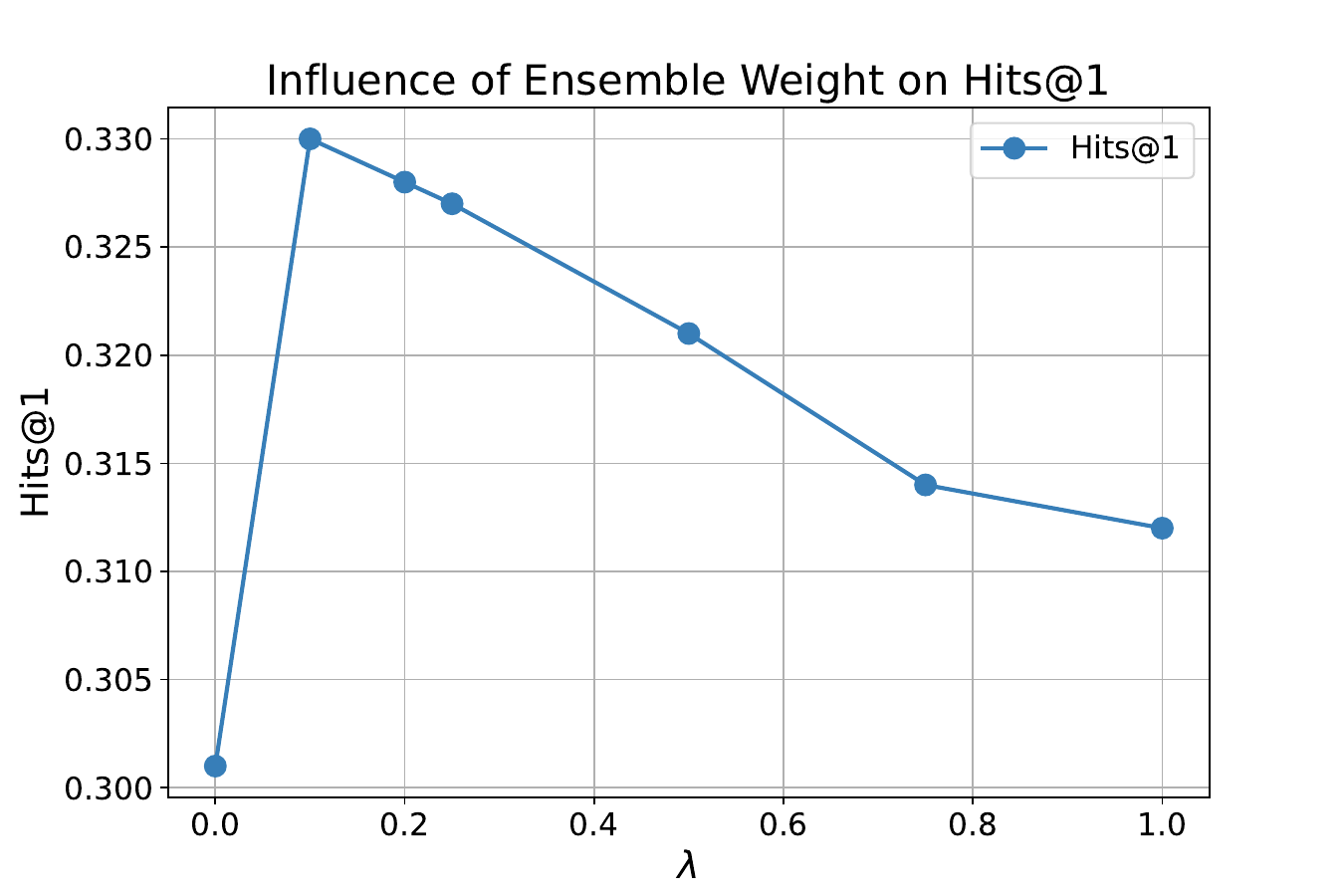}
    }
    \caption{Performance of ONSEP in terms of Hit@1 across various DHAG ensemble weights \(\lambda\) of DHAG. The underlying LLM is InternLM2-7B, processing input histories of length 100. \(\lambda\) represents the weight given to long-term causal event chains. This illustrates how varying \(\lambda\) influences the integration of short-term and long-term reasoning contexts within ONSEP.}
    \label{fig:ensemble_weight}
\end{figure}

\subsection{Performance Comparison of Different Model Scale and Series}

Our analysis reveals that model performance improves with increased parameter scale, with the 20B models outperforming the 7B models, aligning with the scaling law. However, the performance gain from increasing the model size from 7B to 20B is less pronounced and comes with higher computational costs. The ONSEP method enhances performance across different model scales and series, demonstrating its adaptability and effectiveness. Detailed comparisons across model series indicate that ONSEP's improvements are consistent, with InternLM2-7b showing the most significant gains. Further in-depth analysis and discussions are provided in Appendix \ref{Appendix:model_scale_series}.

\subsection{Hyperparameter Sensitivity Analysis}

Our hyperparameter sensitivity analysis highlights the critical role of selecting the right history length \(L\), rule selection thresholds \(k\), and fusion ratios \(\alpha\) for causal rule confidence scores. Finding the right balance between the length of historical input and computational efficiency is essential for optimal model performance. Similarly, precise calibration of rule selection thresholds and fusion ratios is vital for the effective application and updating of causal rules, taking into account factors such as the smoothing factor \(\theta\) and growth factor \(\beta\). These parameters collectively influence the model's ability to adapt and perform accurately over time. A more detailed discussion on these findings and their implications is provided in the Appendix \ref{Appendix:hyperparameter}.

\section{Related Work}
\subsection{Temporal Knowledge Graph Forecasting}

In TKGF, traditional embedding-based methods \cite{jin2019recurrent, zhu2021learning, li2021temporal, han2020xerte, sun2021timetraveler, li2022tirgn} learn representations of the quadruple, showing efficacy in supervised learning. Recent efforts explore LLMs for event prediction. For example, Xu et al. \cite{xu2023pre} use a masking strategy to make the forecasting task similar to predicting missing words. LAMP \cite{shi2023language} utilizes LLMs as a cause generator to reorder candidate outcomes, while ICL has been applied in TKGF \cite{lee2023temporal}, transforming forecasting into a sequence generation task. Besides, GenTKG \cite{liao2023gentkg} leverages few-shot parameter-efficient instruction tuning of LLMs using the training set to enhance inference capabilities. These approaches hold promise for improved generalization and contextual understanding. However, their effectiveness in dynamic real-world scenarios warrants further investigation.

On the other hand, rule-based approaches like Tlogic \cite{liu2022tlogic} focus on interpretability and generalization by learning symbolic rules from TKGs. Despite their strengths, these methods struggle with large search spaces, can't incorporate textual semantics of relations, and rely on static rule sets, limiting their adaptability.

\subsection{Retrieve Augmented Generation}

Retrieval-augmented generation (RAG) significantly enhances LLMs by integrating dynamic external knowledge retrieval \cite{lewis2020retrieval}, mitigating common challenges like hallucinations and slow information updates. Advancements like REPLUG \cite{shi2023replug} and AAR \cite{yu2023augmentation} further enhance RAG. REPLUG refines retrieval models with supervised feedback from language models. AAR, on the other hand, is a versatile plug-in trained with a single source LLM but adaptable to various target LLMs. These adaptive mechanisms of RAG could be particularly beneficial in LLM-based TKGF, improving semantic alignment between queries and retrieved history context.

\subsection{Self-Improving on LLMs}

Researchers have recently proposed methods by using LLM's inherent knowledge as an external database to let LLMs self-improve without annotated datasets and parameter updates. Frameworks like ExpNote \cite{sun2023expnote}, HtT \cite{zhu2023large}, and MoT \cite{li2023mot} facilitate learning from experience, rule induction, and high-confidence thought generation. However, their application to tasks with temporal dimensions remains to be explored. Following this idea, we proposed an ONSEP framework to enable the model to induce rules in real-time during the testing process and then use them for future predictions.

\section{Conclusion}

In this paper, we introduce a novel online neural-symbolic framework, ONSEP, that integrates LLMs with TKGs to achieve adaptive and precise event forecasting in a dynamic online environment. To overcome the challenges of not utilizing experience during testing and relying on a single short-term history, which limits adaptation to new data, we propose a dynamic causal rule mining module and a dual history augmented generation module within the ONSEP framework. This design allows LLMs to access the most recent history to identify patterns, as well as causal relationships from a broader range of past events. Extensive experiments conducted on three benchmark datasets have proven the efficacy of ONSEP in TKGF, surpassing previous methods and demonstrating its broad applicability across diverse LLMs. Our framework shows great potential for future applications in financial forecasting, public sentiment monitoring, and recommendation systems.

\section*{Limitations}
Our method requires multiple uses of large models, leading to increased inference time (see Appendix \ref{Appendix:inference_time}) and higher computational costs compared to simpler models. The effectiveness of the model is influenced by the length of the input context; longer contexts are only useful for LLMs designed to handle them. The method also faces interpretability challenges due to unclear reasoning paths, which can be particularly evident in applications like campaign strategy analysis with the ICEWS dataset. 

Additionally, the method's potential to improve performance is limited for data that lacks a rich semantic understanding or detailed relationships needed to extract comprehensive causal rules. Since there is an upper limit on the length, the model inference uses indexing for the cue words and does not use the lexical setting, ignoring the effect of entity semantics on the results.

\section*{Ethics Statement}
This paper presents a novel online neural-symbolic framework for event prediction, particularly tailored for dynamic real-world environments like TKGF. All experiments are conducted on publicly available datasets. Thus there is no data privacy concern. Meanwhile, this paper does not involve human annotations, and there are no related ethical concerns.

\section*{Acknowledgements}
This work was supported in part by the National Key Research and
Development Program of China under Grant 2022YFB3304602 and the
National Nature Science Foundation of China under Grant 62003344.

\bibliography{custom}
\bibliographystyle{acl_natbib}

\appendix

\section{Algorithm of Semantic-Driven Rule Learning}
\label{Appendix:semantic_driven_rule_learning}

In this section we introduce the Semantic-Driven Rule Learning algorithm in DCRM. The pseudo code can be found at algorithm \ref{alg:semantic_driven_rule_learning}.

\begin{algorithm}
\caption{Semantic-Driven Rule Learning}
\label{alg:semantic_driven_rule_learning}
\begin{algorithmic}[1]
\REQUIRE Historical context $\mathcal{H}_n$ at time $t_n$, Query $(S_q, R_e, ?, t_{n})$, target $O_y$
\ENSURE Causal Rules $\mathcal{CRB}^n(R_e)$

\STATE \textbf{Initialize:} $\mathcal{H}_c \gets \emptyset$, $\mathcal{CRB}^n(R_e) \gets \mathcal{CRB}^{n-1}(R_e)$, $supp \gets \emptyset$, $cove \gets \emptyset$

\FOR{each event $(S_q, r_{c_i}, o_i, t_i)$ in $\mathcal{H}_n$}
    \IF{$o_i$ aligns with $O_y$}
        \STATE $\mathcal{H}_c \gets \mathcal{H}_c \cup \{(S_q, r_{c_i}, o_i, t_i)\}$
        \STATE $supp[r_{c_i}] \gets supp[r_{c_i}] + 1$
    \ENDIF
\ENDFOR

\FOR{each $r_{c_i}$ in $\mathcal{H}_c$}
    \STATE $p_i \gets$ LLM-Causality-Assessment$(r_{c_i}, R_e)$ 
    \STATE Compute $cove_i \gets \frac{supp[r_{c_i}]}{|\mathcal{H}_c|}$
\ENDFOR
\STATE $\mathcal{H}_{c_{topk}} \gets$ Select top-$k$ $r_{c_i}$ based on $p_i$

\FOR{each $r_{c_i}$ in $\mathcal{H}_{c_{topk}}$}
    \STATE Compute $conf_i^n \gets \alpha \cdot p_i + (1 - \alpha) \cdot cove_i$
    \STATE $CR^n(R_e, r_{c_i}) \gets (X, R_e, Y, T_2) \Leftarrow (X, r_{c_i}, Y, T_1)$
    \STATE $\mathcal{CRB}^n(R_e) \gets \mathcal{CRB}^n(R_e) \cup CR^n(R_e, r_{c_i})$
\ENDFOR
\RETURN $\mathcal{CRB}^n(R_e)$

\end{algorithmic}
\end{algorithm}

\label{alg:semantic_driven_rule_learning}

\section{Experimental Details}
\label{Appendix:experimental_details}

In the experiment, we used the ICEWS data sets, including ICEWS14, ICEWS05-15 and ICEWS18, we use the training and validation sets to build historical graphs, which then serve to test our model's predictive accuracy. The specific parameters are shown in the table \ref{table:dataset_details}.

\subsection{Datasets}
\begin{table}[ht]
\centering
\resizebox{\columnwidth}{!}{%
\begin{tabular}{lcccccc}
\toprule 
\textbf{Dataset} & \textbf{\# Ents} & \textbf{\# Rels} & \textbf{Train} & \textbf{Valid} & \textbf{Test} & \textbf{Interval} \\
\midrule
ICEWS14     & 7,128  & 230 & 74,845 & 8,514 & 7,371 & 24 hours\\
ICEWS05-15  & 10,094 & 251 & 368,868 & 46,302 & 46,159 & 24 hours\\
ICEWS18     & 23,033 & 256 & 373,018 & 45,995 & 49,545 & 24 hours\\
\bottomrule
\end{tabular}%
}
\caption{Statistics of the datasets.}
\label{table:dataset_details}
\end{table}

ICEWS14 and ICEWS18 have significant differences in data distribution. Specifically, the number of entities in ICEWS18 far exceeds that in ICEWS14. This can be clearly demonstrated in the dataset statistics:

\begin{table}[ht]
\centering
\resizebox{0.9\columnwidth}{!}{%
    \begin{tabular}{lccc}
    \toprule 
    \textbf{Dataset} & \textbf{Entities} & \textbf{Relations} & \textbf{Time Span} \\
    \midrule
    ICEWS14 & 7,128    & 230       & 2014      \\
    ICEWS18 & 23,033   & 256       & 2018      \\
    Overlap & 4,685    & 226       & -         \\
    Overlap (\%) & 20.3\% & 88.3\%  & - \\
    \bottomrule
    \end{tabular}
    }
    \caption{Overlap statistics between ICEWS14 and ICEWS18. The "Overlap (\%)" shows the percentage of entities and relations in ICEWS18 that overlap with ICEWS14.}
    \label{table:overlap_statistics}
\end{table}

\subsection{Evaluation Metrics}

 To assess ONSEP's ability to rank candidates for event prediction without recalling all entities, we use link prediction metrics like Hit@k (where k=1, 3, 10). Hit@k is an evaluation metric that measures how often the model correctly places entities within the top k positions of the ranking list for a given query. Our evaluation specifically focuses on a time-aware setting, where we apply filtering to remove valid candidates not pertinent to the specific query.

\subsection{Baselines}
All large models used in the paper's experiments are shown here, and the series and parameters are reported, as shown in Table \ref{table:LLMs_used}.

\begin{table}[H]
\centering
\resizebox{\columnwidth}{!}{%
\begin{tabular}{@{}llc@{}}
\toprule
\textbf{Model Family} & \textbf{Model Name} & \textbf{\# Params} \\
\midrule
Qwen \cite{bai2023qwen} & qwen-7b & 7B \\
Baichuan2 \cite{yang2023baichuan} & baichuan2-7b & 7B \\
LLaMA2 \cite{touvron2023llama} & llama-2-7b & 7B \\
InternLM2 \cite{team2023internlm} & internlm2-7b & 7B \\
& internlm2-20b & 20B \\
GPT-NeoX \cite{black2022gpt} & gpt-neox-20b & 20B \\
\bottomrule
\end{tabular}
}
\caption{LLMs Used in the Study.}
\label{table:LLMs_used}
\end{table}

\subsection{Implementation Details}
\label{Implementation}

The experimental setup for the ICL method aligns with that of TKG-ICL \cite{lee2023temporal}, except for a change in the model. The input provided to the model is based on indexing rather than lexical content. For the 20B model, due to computational resource constraints, we employed a 4-bit quantization approach for loading the model and performing inference. All experiments were conducted on GeForce RTX 3090 GPU with 24GB Memory. 


\section{Case Study}
\label{Appendix:case_study}

This section delves into the ICEWS14 case study, demonstrating ONSEP's innovative approach in event prediction. The aim is to showcase the method's operational dynamics and its practical relevance.

\subsection{Operational Setup and Input Scenarios}
\begin{figure}[H]
    \centering
    \hspace*{-1.2cm} 
    \begin{adjustbox}{max width=1.25\columnwidth}
        \includegraphics[width=1.25\columnwidth]{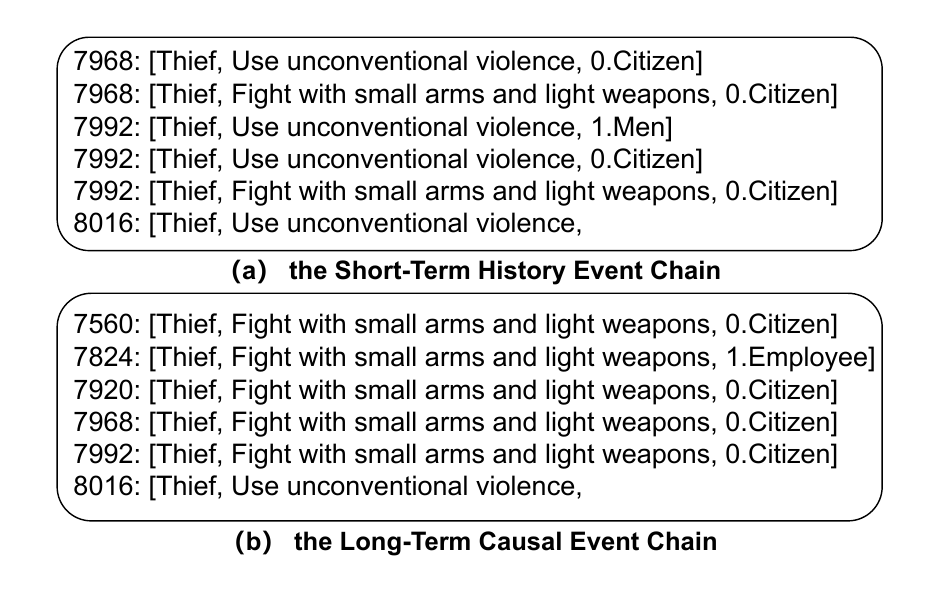}
    \end{adjustbox}
     \caption{Example of short-term and long-term historical event chains used by ONSEP.}
    \label{his_event_chain}
\end{figure}

Initially, consider an input history length limit of 5 events. For the query scenario involving a hypothetical entity \colorbox{lightgray}{$\texttt{Thief}$} engaged in \colorbox{lightgray}{$\texttt{Use\ unconventional\ violence}$}, ONSEP's task is to predict potential outcomes based on past events. The ICL method, serving as a comparative baseline, identifies only a short-term history sequence. In contrast, ONSEP extends the analysis both short-term and long-term historical data as Figure \ref{his_event_chain} shows.

\subsection{Causal Analysis and Predictive Accuracy}
ONSEP employs learned causal rules to enhance prediction accuracy, for instance, inferring that \colorbox{lightgray}{$\texttt{Use\ unconventional\ violence}$} may evolve from \colorbox{lightgray}{$\texttt{Fight\ with\ small\ arms\ and}$} \colorbox{lightgray}{$\texttt{light\ weapons}$} with a confidence score (\(conf\): 0.76). This analysis is dynamically derived from both the LLM's assessment and observed data frequencies.

When multiple rules apply, ONSEP retrieves relevant causative events from the historical graph for each rule and contextualizes them chronologically. This process leverages direct and precise causal relationships and enables ONSEP to integrate a broader range of historical insights, extending further back in time than baseline models. As the temporal dataset grows, ONSEP's rule repository continuously evolves, improving the accuracy of historical context retrieval and updating the confidence levels of relational rules.

\section{Additional Experiment Results}

\subsection{Performance Comparison in Different Model Series}
\label{Appendix:model_scale_series}
\paragraph{Model Scale}

To analyze the impact of the model parameter scale on performance, we selected the 7B and 20B models from the InternLM2 series as the foundation models for ONSEP, with an input length of 100. As shown in Table \ref{tab:scale}, the 20B model performs better than the 7B model on the ICL method, consistent with our expectations and in line with the scaling law. The ONSEP method demonstrates significant improvements across both model scales, but the growth on the 20B model is relatively lower than on the 7B, and it is more time-consuming. The improvements in Hit@1 are 9.63\% and 2.45\% for the 7B and 20B models, respectively. This also proves that larger parameter LLMs have advantages in feature capturing, enabling better reasoning performance, but it's important to consider the balance between performance gains and computational costs.

\begin{table}[htbp]
\centering
\resizebox{\columnwidth}{!}{%
    \label{tab:scale}
    \begin{tabular}{@{}lllll|l@{}}
    \toprule
    Model & Hit@1 & Hit@3 & Hit@10 & & Time \\ \midrule
    \textbf{InternLM2-7b-ICL}   & 0.301 & 0.432 & 0.560  & & - \\
    \textbf{InternLM2-7b-ONSEP}& 0.330 & 0.464 & 0.570 & & 2 h 32 min \\
    \rowcolor{gray!30} \textit{$\Delta$ Improve} & 9.63\% & 7.41\% & 1.79\% & & \\
    \textbf{InternLM2-20b-ICL} & 0.326 & 0.455 & 0.57  & & - \\
    \textbf{InternLM2-20b-ONSEP}& \textbf{0.334} & \textbf{0.467} & \textbf{0.571} & & 7h 10 min\\
    \rowcolor{gray!30} \textit{$\Delta$ Improve} & 2.45\% & 2.64\% & 0.18\% & & \\
    \bottomrule
    \end{tabular}
}
\caption{Comparative Analysis of Performance Enhancement Across Varied Model Sizes.}
\label{tab:scale}
\end{table}

\paragraph{Model Series}

\begin{figure}[t]
    \centering
    \begin{adjustbox}{max width=\columnwidth} 
    \includegraphics[width=1.0\textwidth]{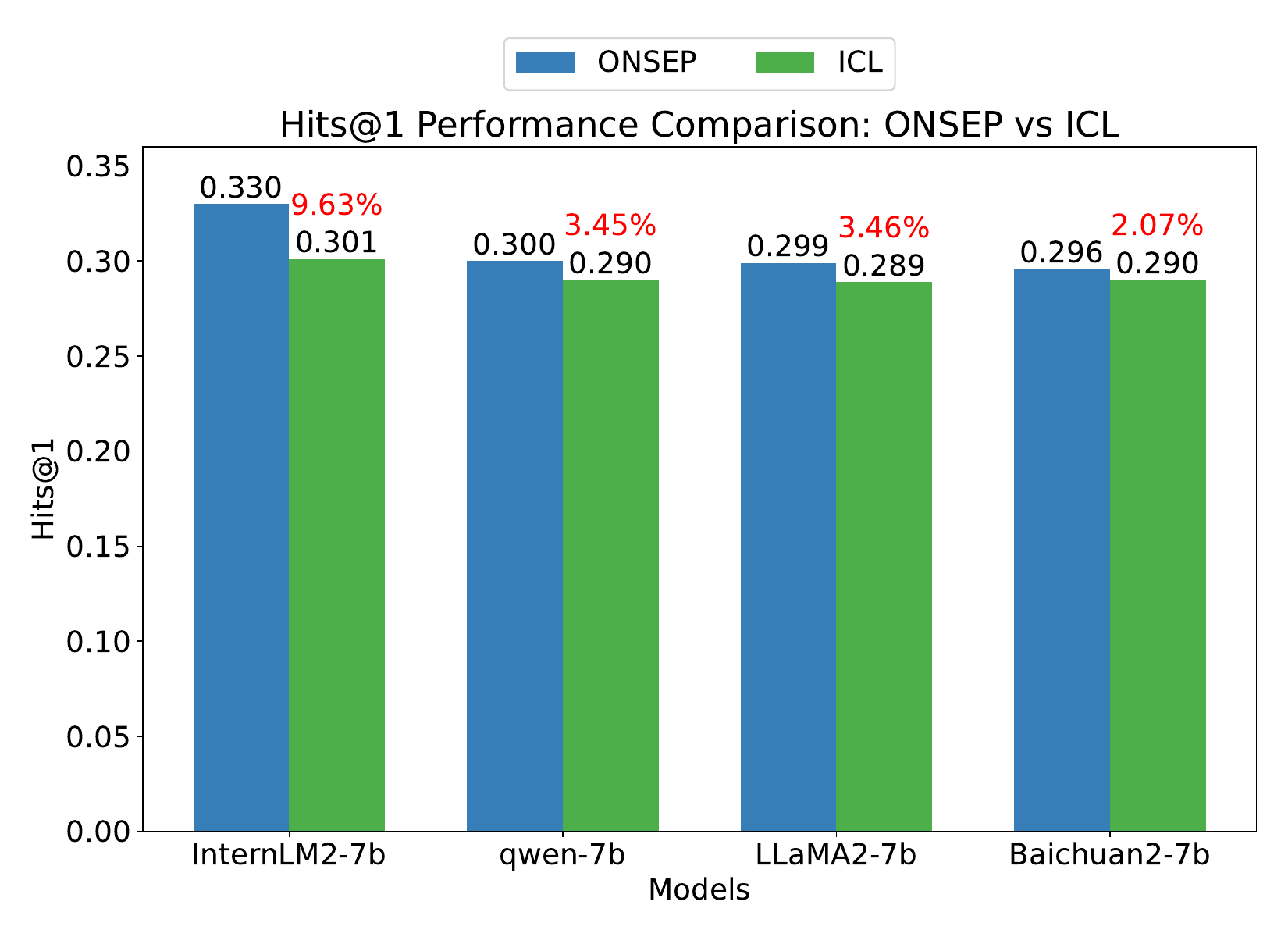}
    \end{adjustbox}
    \caption{Performance comparison across various model series under ONSEP and ICL methods with the percentage improvement indicated in red above the green bars.}
    \label{fig:ONSEP icl}
\end{figure}

To compare the performance of different models under two scenarios and demonstrate ONSEP's improvement over ICL, we use 7B models from the InternLM2, Qwen, LLaMA2, and Baichuan2 series. Figure \ref{fig:ONSEP icl} compares the performance of each model under ONSEP and ICL conditions. The comparison reveals that InternLM2-7b outperforms others with ICL and shows the most significant improvement with ONSEP, leading among models of similar size. While Qwen, LLaMA2, and Baichuan2 have similar performances with ICL, LLaMA2 and Qwen exhibit greater improvements with ONSEP than Baichuan2.

For the 20B models, we examine GPT-NeoX and InternLM2, as indicated in Table \ref{tab:model_improvement2}.

The performance differences may stem from variations in vocabulary, tokenization, training data, decoding strategies, and BPE encoding specifics \cite{sennrich2016neural} among the different model series. InternLM2, with its innovative pre-training and optimization, excels in long-context tasks by capturing long-term dependencies. InternLM2's superior performance may be due to a progressive training approach, starting with 4k tokens and extending to 32k tokens. ONSEP consistently enhances the performance across these models, showcasing the methodology's generalizability. Additional metrics and detailed data are available in Appendix \ref{Appendix:compare_series}.

\begin{table}[ht]
\centering
\resizebox{\columnwidth}{!}{%
    \begin{tabular}{@{}ccccc@{}}
    \toprule
    Model & Hit@1 & Hit@3 & Hit@10 \\ \midrule
    \textbf{GPT-NeoX-20b-ICL} & 0.314 & 0.446 & 0.560 \\
    \textbf{GPT-NeoX-20b-ONSEP} & 0.320 & 0.454 & 0.563 \\
    \rowcolor{gray!30} \textit{$\Delta$ Improve} & 1.91\% & 1.79\% & 0.54\% \\
    \textbf{InternLM2-20b-ICL} & 0.326 & 0.455 & 0.57 \\
    \textbf{InternLM2-20b-ONSEP} & 0.334 & 0.467 & 0.571 \\
    \rowcolor{gray!30} \textit{$\Delta$ Improve} & 2.45\% & 2.64\% & 0.18\% \\
    \bottomrule
    \end{tabular}
}
\caption{Comparison of performance across two 20B parameter model series, highlighting the percentage improvement ONSEP achieves over ICL.}
\label{tab:model_improvement2}
\end{table}

Table \ref{tab:model_improvement} shows the ICL methods for four different series of models of comparable size and the proposed ONSEP method. It is observed that InternLM2 performs the best in terms of various indicators and improvements. For Hit@1 and Hit@3, the performances of the other three models are similar, ranked in the order of Hit@1 as Qwen, Baichuan2, and Llama2. In terms of the improvement in Hit@10, the method achieved a 13.41\% increase on Llama2. The pre-training data distribution and tokenization methods of these different open-source models vary, which impacts the performance.

\label{Appendix:compare_series}
\begin{table}[ht]
\centering
\resizebox{0.8\columnwidth}{!}{%
    \begin{tabular}{@{}ccccc@{}}
    \toprule
    Model & Hit@1 & Hit@3 & Hit@10 \\ \midrule
    \textbf{InternLM2-7b-ICL} & 0.301 & 0.432 & 0.560 \\
    \textbf{InternLM2-7b-ONSEP} & 0.330 & 0.464 & 0.570 \\
    \rowcolor{gray!30} \textit{$\Delta$ Improve} & 9.63\% & 7.41\% & 1.79\% \\
    \textbf{Qwen-7b-ICL} & 0.290 & 0.421 & 0.530 \\
    \textbf{Qwen-7b-ONSEP} & 0.300 & 0.440 & 0.560 \\
    \rowcolor{gray!30} \textit{$\Delta$ Improve} & 3.4\% & 4.51\% & 5.66\% \\
    \textbf{LLama2-7b-ICL} & 0.289 & 0.412 & 0.440 \\
    \textbf{LLama2-7b-ONSEP} & 0.299 & 0.438 & 0.499 \\
    \rowcolor{gray!30} \textit{$\Delta$ Improve} & 3.5\% & 6.31\% & 13.41\% \\
    \textbf{Baichuan2-7b-ICL} & 0.290 & 0.416 & 0.530 \\
    \textbf{Baichuan2-7b-ONSEP} & 0.296 & 0.437 & 0.552 \\
    \rowcolor{gray!30} \textit{$\Delta$ Improve} & 2.06\% & 5.05\% & 4.15\% \\
    \bottomrule
    \end{tabular}
}
\caption{Performance Comparison and Improvement Across Models.}
\label{tab:model_improvement}
\end{table}

\subsection{Ensemble Weight in DHAG}
\label{Appendix:ensemble_weight}

Table \ref{tab:ensemble_weight} demonstrates the effect of adjusting the ensemble weight $\lambda$ within the Dual History Augmented Generation (DHAG) of the ONSEP model on performance metrics Hit@1, Hit@3, and Hit@10. The data reveals that the optimal result for Hit@1 is achieved at $\lambda=0.1$, while the highest coverage rate for Hit@10 is observed at $\lambda=0.5$, indicating improvements in both metrics. Extreme values of $\lambda$ (degenerating to the ICL method) lead to suboptimal outcomes, highlighting the dual context's ability to enhance both accuracy and coverage, as well as the importance of balancing short-term and long-term historical contexts.

As $\lambda$ increases, accuracy decreases, while the answer coverage remains constant, suggesting that DHAG enhances model performance by slightly increasing the probability of selecting the target (Hit@1).

\begin{table}[ht]
\centering
\
\resizebox{0.8\columnwidth}{!}{%
\begin{tabular}{c|ccc}
\toprule
\textbf{Ensemble Weight $\lambda$ }& \textbf{Hit@1} & \textbf{Hit@3} & \textbf{Hit@10} \\
\midrule
0 & 0.301 & 0.432 & 0.560 \\
\midrule
0.1 & 0.330 & 0.464 & 0.570 \\
\midrule
0.2 & 0.328 & 0.462 & 0.571 \\
\midrule
0.25 & 0.327 & 0.462 & 0.572 \\
\midrule
0.5 & 0.321 & 0.460 & 0.573 \\
\midrule
0.75 & 0.314 & 0.456 & 0.572 \\
\midrule
1 & 0.312 & 0.445 & 0.552 \\
\bottomrule
\end{tabular}
}
\caption{Results for different choices of ensemble weight of DHAG. The LLM based in ONSEP is InternLM2-7B with a history length of 100.}
\label{tab:ensemble_weight}
\end{table}

\section{Hyperparameter Sensitivity Analysis}
\label{Appendix:hyperparameter}

This section looks into how different hyperparameters affect our method. We use the ICEWS14 dataset and the InternLM2-7B model for this analysis, but we find similar patterns in other datasets too.

The set of tested hyperparameter ranges and best parameter values for ONSEP are displayed in Table \ref{tab:hyper_overview}. The best hyperparameter values are chosen based on the Hit@1.

\begin{table}[ht]
\centering
\resizebox{1.0\columnwidth}{!}{%
\begin{tabular}{c|cc}
\toprule
\textbf{Hyperparameter} & \textbf{Set} & \textbf{Best} \\
\midrule
Ensemble Weight $\lambda$ & \{0, 0.1, 0.2, 0.25, 0.5, 0.75, 1\}& 0.1\\
\midrule
History Length $L$ & \{10, 30, 50, 100, 150, 200\} & 200 \\
\midrule
Select rules num $k$ & \{0, 1, 3, 5, 10, 20\} & 20\\
\midrule
Causality Ratio $\alpha$& \{0,  0.1,  0.2,  0.25, 0.5,  0.75, 1\} & 0.1\\
\midrule
Smooth Factor $\theta$ & \{0, 0.25, 0.5, 0.75, 1\} & 0.25\\
\midrule
Growth Factor $\beta$ & \{0, 0.1, 0.15, 0.2, 0.25, 0.3, 0.5, 0.75, 1\} & 0.2\\
\bottomrule
\end{tabular}
}
\caption{Overview of hyperparameters.}
\label{tab:hyper_overview}
\end{table}

\subsection{History Length}

\begin{figure}[t]
    \centering
    \begin{adjustbox}{max width=\columnwidth} 
    \includegraphics[width=1.0\textwidth]{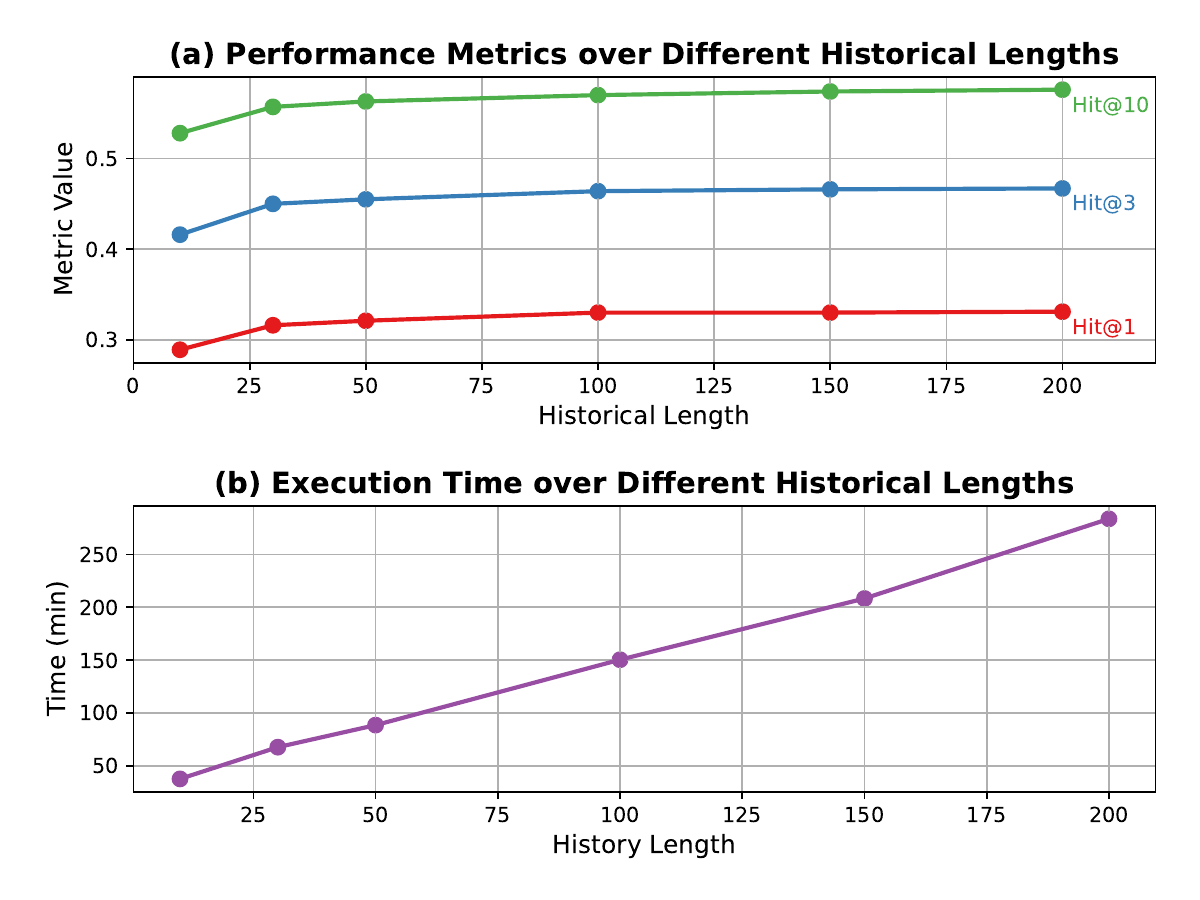}
    \end{adjustbox}
    \caption{How History Length Affects Performance and Prediction Time.}
    \label{fig:his_len}
\end{figure}

We assess the impact of varying historical lengths on performance metrics (Hit@N) and the model's inference time, as shown in Figure \ref{fig:his_len}. Increasing historical length generally leads to better performance but also longer inference times. The most significant improvements occur up to a history length of 100; after that, the benefits level off, indicating an optimal balance between performance gains and computational efficiency is necessary.

Table \ref{tab:his_len} demonstrates the impact of different history lengths $L$ on model performance. As the history length increases, all performance metrics (Hit@1, Hit@3, and Hit@10) improve, but the growth saturates with longer $L$. If computational resources are limited or fast inference is required, $L$ should be appropriately reduced to balance performance and efficiency.

\begin{table}[ht]
\centering
\resizebox{0.8\columnwidth}{!}{%
\begin{tabular}{c|ccc}
\toprule
\textbf{History Length $L$} & \textbf{Hit@1} & \textbf{Hit@3} & \textbf{Hit@10} \\
\midrule
10 & 0.289 & 0.416 & 0.528 \\
\midrule
30 & 0.316 & 0.450 & 0.557 \\
\midrule
50 & 0.321 & 0.455 & 0.563 \\
\midrule
100 & 0.330 & 0.464 & 0.570 \\
\midrule
150 & 0.330 & 0.466 & 0.574 \\
\midrule
200 & 0.331 & 0.467 & 0.576 \\
\bottomrule
\end{tabular}
}
\caption{Results for different choices of history length $L$.}
\label{tab:his_len}
\end{table}

\subsection{Maximum Number of Rules Selected}

In the DCRM module, we experiment with changing the upper limit, K, of optimal rules selected after causality evaluation by the LLM. This parameter shows how selective we are in filtering causality semantics and influences the causal rule set size. As K increases, Hit@N performance improves, up to a certain point. Setting K to infinity (meaning no selection) causes a slight drop in performance from the optimum, suggesting that a larger rule set helps, but keeping high causality confidence is essential to minimize noise and avoid rule application disruption.

The results of selecting the maximum number of rules $k$, as shown in \ref{tab:select rules num}, indicate that the model's performance increases with the number of rules selected, reaching a saturation point. The last row suggests that not selecting rules after LLM evaluation results in slightly lower performance, highlighting the importance of optimal rule selection for achieving peak model effectiveness.

\begin{table}[ht]
\centering
\resizebox{0.8\columnwidth}{!}{%
\begin{tabular}{c|ccc}
\toprule
\textbf{Select rules num $k$} & \textbf{Hit@1} & \textbf{Hit@3} & \textbf{Hit@10} \\
\midrule
0 & 0.318 & 0.454 & 0.561 \\
\midrule
1 & 0.327 & 0.461 & 0.569 \\
\midrule
3 & 0.328 & 0.462 & 0.570 \\
\midrule
5 & 0.329 & 0.462 & 0.570 \\
\midrule
10 & 0.330 & 0.464 & 0.570 \\
\midrule
20 & 0.329 & 0.462 & 0.570 \\
\bottomrule
\end{tabular}
}
\caption{Results for different choices of Select rules num $k$.}
\label{tab:select rules num}
\end{table}

\subsection{Fusion Ratio of Causal Rule Confidence Scores}

We experiment with various fusion ratios \(\alpha\), mixing large model predicted probabilities \(p\) and coverage \(cove\) from contextual frequencies to see their effect on performance. A balanced \(\alpha\) setting achieves optimal performance, highlighting the importance of both predicted probabilities and contextual coverage in determining causality confidence.

Regarding the fusion ratio of causal rule confidence scores $\alpha$, the table illustrates how adjusting $\alpha$ affects model performance. The data in Table \ref{tab:Causality Assessment Ratio} show that the model performs best at Hit@1 with $\alpha=0.5$ and reaches peak performance at Hit@3 with $\alpha=0.75$. This indicates that an appropriate $\alpha$ value can balance the model's prediction accuracy and coverage range. Too high or too low $\alpha$ values may degrade performance on certain metrics.

\begin{table}[ht]
\centering
\resizebox{0.8\columnwidth}{!}{%
\begin{tabular}{c|ccc}
\toprule
\textbf{Causality Assessment Ratio $\alpha$} & \textbf{Hit@1} & \textbf{Hit@3} & \textbf{Hit@10} \\
\midrule
0 & 0.325 & 0.460 & 0.568 \\
\midrule
0.25 & 0.327 & 0.461 & 0.569 \\
\midrule
0.5 & 0.329 & 0.462 & 0.570 \\
\midrule
0.75 & 0.327 & 0.464 & 0.570 \\
\midrule
1 & 0.323 & 0.461 & 0.566 \\
\bottomrule
\end{tabular}
}
\caption{Results for varying Causality Assessment Ratio $\alpha$.}
\label{tab:Causality Assessment Ratio}
\end{table}

\subsection{Causal Rule Confidence Update Function}
In DCRM's dynamic update process, adjusting causal rule confidence through factors like smoothing and growth is crucial for adapting to new data and trends. A balanced smoothing factor $\theta$ setting works well to incorporate both historical and new data, with larger values improving performance on the ICEWS dataset, suggesting stability is needed in dynamic data distributions. On the other hand, a careful setting of the growth factor $\beta$ improves performance. It shows the importance of slowly increasing focus on rules that have worked in the past without making their confidence scores too high too quickly.

\paragraph{smooth factor}

The results for the smooth factor $\theta$ in the causal rule confidence update function ( \ref{tab:smooth_factor_results}) suggest that a moderate $\theta$ value balances historical and new information effectively. With a smooth factor of 0, the confidence scores are updated solely based on new samples, while a factor of 1 retains the initial confidence assessments. A slightly higher $\theta$ may be preferable for stable rule adaptation in dynamic data scenarios, whereas a lower value could be suitable if there are significant intervals between samples.

\begin{table}[ht]
\centering
\resizebox{0.8\columnwidth}{!}{%
\begin{tabular}{c|ccc}
\toprule
\textbf{Smooth Factor $\theta$} & \textbf{Hit@1} & \textbf{Hit@3} & \textbf{Hit@10} \\
\midrule
0 & 0.325 & 0.462 & 0.570 \\
\midrule
0.25 & 0.330 & 0.464 & 0.570 \\
\midrule
0.5 & 0.327 & 0.463 & 0.570 \\
\midrule
0.75 & 0.327 & 0.462 & 0.571 \\
\midrule
1 & 0.324 & 0.461 & 0.566 \\
\bottomrule
\end{tabular}
}
\caption{Results for different smooth factor $\theta$ in the confidence score function.}
\label{tab:smooth_factor_results}
\end{table}

\paragraph{growth factor}

Table \ref{tab:growth_factor_results} shows that appropriately setting the growth factor $\beta$ can optimize model performance without being excessively high. A cautious, lower growth factor can incrementally increase attention to historically effective rules, enhancing their confidence scores. Conversely, too high a $\beta$ may introduce more noise, adversely affecting performance.

\begin{table}[H]
\centering
\resizebox{0.8\columnwidth}{!}{%
\begin{tabular}{c|ccc}
\toprule
\textbf{Growth Factor $\beta$} & \textbf{Hit@1} & \textbf{Hit@3} & \textbf{Hit@10} \\
\midrule
0 & 0.328 & 0.465 & 0.571 \\\midrule
0.1 & 0.328 & 0.464 & 0.570 \\\midrule
0.15 & 0.328 & 0.464 & 0.571 \\\midrule
0.2 & 0.330 & 0.464 & 0.570 \\\midrule
0.25 & 0.325 & 0.461 & 0.568 \\\midrule
0.3 & 0.325 & 0.461 & 0.568 \\\midrule
0.5 & 0.324 & 0.458 & 0.565 \\\midrule
0.75 & 0.320 & 0.457 & 0.567 \\\midrule
1 & 0.323 & 0.457 & 0.566 \\
\bottomrule
\end{tabular}
}
\caption{Results for different growth factor $\beta$ in the confidence score function.}
\label{tab:growth_factor_results}
\end{table}

\section{Inference Time}
\label{Appendix:inference_time}
Our tests on an RTX 3090 showed that ONSEP's inference time (see Table \ref{table:inference_time}) is about twice that of ICL due to multiple LLM calls, but it's still suitable for scenarios like the ICEWS dataset where immediate real-time responses aren't crucial.

\begin{table}[H]
\centering
\resizebox{1.0 \columnwidth}{!}{%
\begin{tabular}{c|ccc}
\toprule
\textbf{Model} & \textbf{ICEWS14} & \textbf{ICEWS05-15} & \textbf{ICEWS18} \\
\midrule
InternLM2-7B-ICL (L=100) & 4.16 it/s & 3.14 it/s & 3.37 it/s \\
\midrule
InternLM2-7B-ONSEP (L=100) & 1.94 it/s & 1.47 it/s & 1.66 it/s \\
\bottomrule
\end{tabular}
}
\caption{Inference times comparison for different methods on ICEWS datasets.}
\label{table:inference_time}
\end{table}

\end{document}